\documentclass[sigconf]{acmart}

\AtBeginDocument{%
  \providecommand\BibTeX{{%
    \normalfont B\kern-0.5em{\scshape i\kern-0.25em b}\kern-0.8em\TeX}}}

\usepackage{amsmath}
\usepackage{bbm}
\usepackage{graphicx}
\usepackage{multirow}
\usepackage{subcaption}
\usepackage[linesnumbered,vlined,ruled]{algorithm2e}
\usepackage{xspace}
\usepackage{makecell}
\newcommand{\alg}{SchemaEGC\xspace}
\newcommand{\xhdr}[1]{\vspace{2mm}{\noindent\bfseries #1}.}

\usepackage{xparse}
\NewDocumentCommand{\heng}
{ mO{} }{\textcolor{red}{\textsuperscript{\textit{Heng}}\textsf{\textbf{\small[#1]}}}}
\NewDocumentCommand{\hh}
{ mO{} }{\textcolor{blue}{\textsuperscript{\textit{HH}}\textsf{\textbf{\small[#1]}}}}
\NewDocumentCommand{\YS}
{ mO{} }{\textcolor{green}{\textsuperscript{\textit{YS}}\textsf{\textbf{\small[#1]}}}}

\setcopyright{none}
\renewcommand\footnotetextcopyrightpermission[1]{}
\begin{document}

\fancyhead{}

\title{Schema-Guided Event Graph Completion}

\author{Hongwei Wang$^{1}$, Zixuan Zhang$^{1}$, Sha Li$^{1}$, Jiawei Han$^{1}$,\\ Yizhou Sun$^{2}$, Hanghang Tong$^{1}$, Joseph P. Olive$^{3}$, Heng Ji$^{1}$\\
\small{$^{1}$University of Illinois Urbana-Champaign $^{2}$University of California, Los Angeles \\
$^{3}$JP Olive S\&T Management LLC} \\
\texttt{\{hongweiw, zixuan11, shal2, hanj, htong, hengji\}@illinois.edu},\texttt{ yzsun@cs.ucla.edu}, \texttt{ joseph.olive.ctr@darpa.mil}\\
}

\begin{abstract}
    We tackle a new task, \textit{event graph completion}, which aims to predict missing event nodes for event graphs.
    Existing link prediction or graph completion methods have difficulty dealing with event graphs, because they are usually designed for a single large graph such as a social network or a knowledge graph, rather than multiple small dynamic event graphs.
    Moreover, they can only predict missing edges rather than missing nodes.
    In this work, we propose to utilize \textit{event schema}, a template that describes the stereotypical structure of event graphs, to address the above issues.
    Our schema-guided event graph completion approach first maps an instance event graph to a subgraph of the schema graph by a heuristic subgraph matching algorithm.
    Then it predicts whether a candidate event node in the schema graph should be added into the instantiated schema subgraph by characterizing two types of local topology of the schema graph: \textit{neighbors} of the candidate node and the subgraph, and \textit{paths} that connect the candidate node and the subgraph.
    These two modules are later combined together for the final prediction.
    We also propose a self-supervised strategy to construct training samples, as well as an inference algorithm that is specifically designed to complete event graphs. 
    Extensive experimental results on four datasets demonstrate that our proposed method achieves state-of-the-art performance, with $4.3\%$ to $19.4\%$ absolute F1 gains over the best baseline method on the four datasets.
\end{abstract}

\maketitle

\section{Introduction}
\begin{figure}[t]
	   \centering
	   \includegraphics[width=0.42\textwidth]{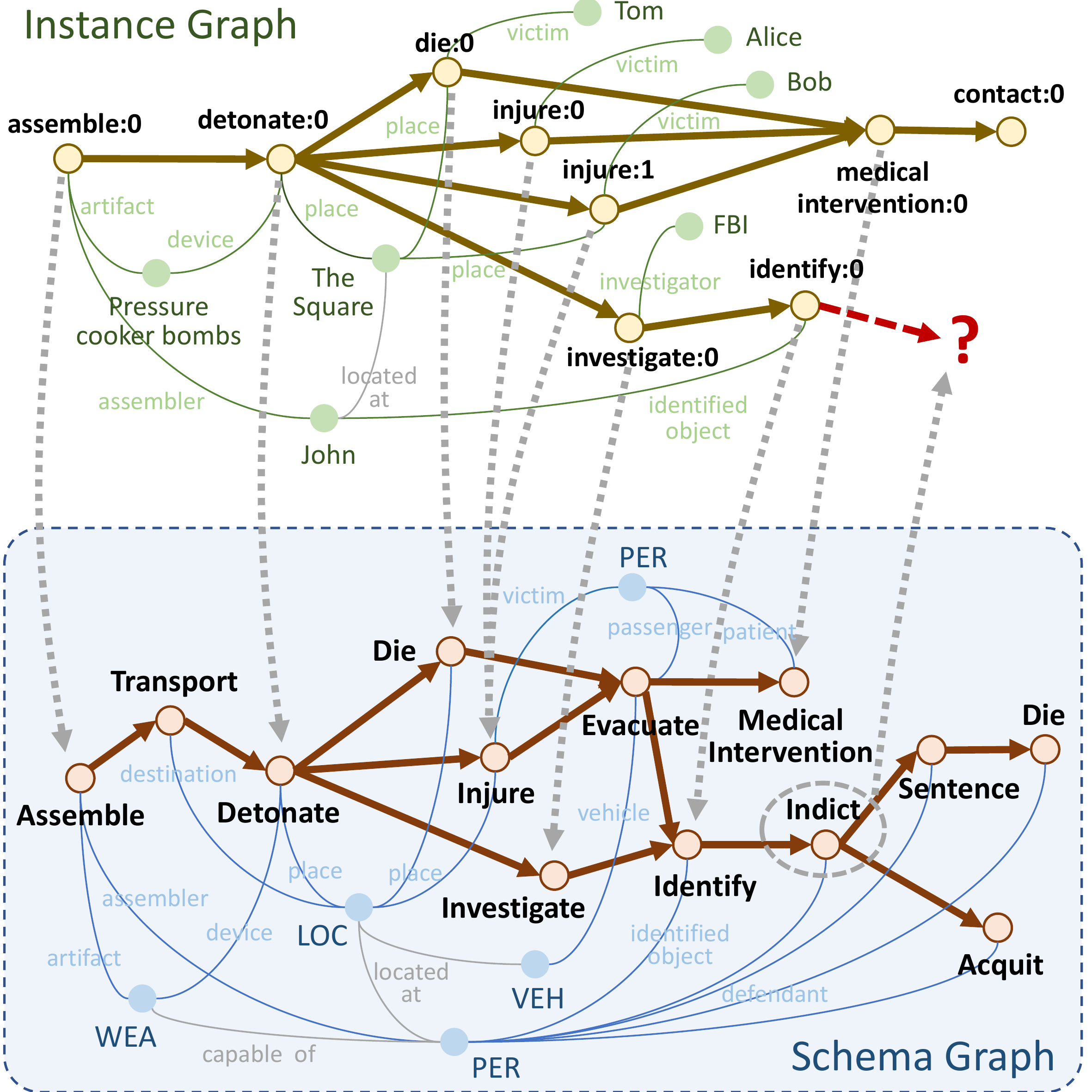}
	   \caption{An illustrative example of schema-guided event graph completion. The above is an (incomplete) instance graph about a bombing event, and the below is the schema of general improvised explosive device (IED) bombing. The grey dotted arrows between the two graphs denote the event node matching relation (entity node matching is omitted for clarity). The schema can inject human knowledge into instance graphs and help predict missing events such as \textsc{Indict}.}
	   \label{fig:example}
	   \vspace{-0.05in}
    \end{figure}

    \textit{Event graphs} \citep{glavavs2015construction} are structured representation of real-world complex events that are automatically extracted from natural language texts such as news.
    An event graph consists of event nodes and entity nodes, as well as their relations including event-event temporal links, event-entity argument links, and entity-entity relations.
    The upper part of Figure \ref{fig:example} gives an example of a complex bombing event extracted from a news report, where yellow and green nodes denote event mentions and entity mentions, respectively. 
    
    Nonetheless, a practical issue on event graphs is that they are often incomplete and noisy, because existing information extraction methods have limited performance to understand ambiguous natural language, and many atomic events are often not explicitly described in the source documents such as news articles. 
    For example, in the instance graph in Figure \ref{fig:example}, the \textsc{injure:0} event probably has an argument \textsc{place} linked to entity \textsc{The Square}, and the next-step event after \textsc{identify:0} is also missing.
    A common solution to this problem is to utilize off-the-shelf algorithms for link prediction \citep{zhang2018link,wang2018shine,lei2019gcn} or graph completion \citep{zhang2019quaternion,goel2020diachronic,wang2021relational} to refine raw event graphs.
    However, they have difficulty dealing with event graphs in the following two aspects:
    (1) Existing link prediction or graph completion methods usually focus on a \textit{single large} graph (e.g., an online social network or knowledge graph with thousands or even millions of nodes), but an event graph dataset consists of \textit{multiple small} instance event graphs, each of which is extracted from a cluster of topically-related news articles  and contains dozens of nodes only.
    The instance event graphs are usually independent (e.g., describing different bombing events) but follow the similar pattern (e.g., all bombing events are similar), which, unfortunately, cannot be characterized by existing methods.
    (2) Existing link prediction or graph completion methods can only detect a missing link between two nodes that already exist in the graph, but they cannot tell if a new node is missing from the graph and how this new node should be connected to existing nodes.
    For example, in the instance graph in Figure \ref{fig:example}, it is unlikely to predict the next-step event after \textsc{identify:0} with the existing link prediction or graph completion methods.
    However, note that a significant difference between event graphs and knowledge graphs is that, knowledge graphs are entity-centric and consist of static entity-entity relations, while event graphs are event-centric and dynamic, where new events and entities may appear as the graph evolves.
    Therefore, predicting missing events is a realistic, informative, and exciting task for event graph completion \cite{li2021future}.
    
    To address the limitations of existing methods when applied to event graphs, a promising solution is to make use of event schemas.
    An \textit{event schema} (a.k.a. complex event template) is a generic and abstract representation of a specific type of complex events that encodes their stereotpyical structure.
    Event schemas can be either generated automatically by machines \cite{granroth2016happens,weber2018event,weber2020causal,li2020connecting,li2021future} from a large collection of historical event graphs, or manually curated by human.
    The lower part of Figure \ref{fig:example} illustrates the schema of general improvised explosive device (IED) bombing, where red and blue nodes represent the types of events and entities, respectively.
    We propose to utilize event schemas to help solve the two aforementioned issues:
    An event schema serves as a template for a particular type of complex events, which represents the generalized knowledge in this particular scenario and enables us to model the common pattern of instance graphs;
    Moreover, an event schema can be seen as a pool of inter-connected candidate events, which provides us with new event nodes that can be added into an incomplete instance event graph (e.g., the \textsc{Indict} node in Figure \ref{fig:example}).
    
    In this paper, we propose a schema-guided approach for event graph completion.
    Given an incomplete instance event graph, we aim to predict whether a candidate event node from the schema graph is missing for the instance graph.
    To build the connection between the schema graph and instance graphs, we propose first using a two-stage heuristic subgraph matching algorithm to map an instance graph to a subgraph of the schema graph (see Figure \ref{fig:example} for example), which can greatly reduce the time overhead of the exact subgraph matching algorithm.
    After the subgraph matching step, our problem is equivalent to inferring whether the candidate node is missing for the matched subgraph of the schema. 
    We therefore explore two types of local topology for a subgraph-node pair within the schema graph:
    (1) \textit{Neighbors}.
    It is important to capture the neighboring node types of a given node/subgraph in the schema graph, because they provide us with valuable information about what the nature of the given node/subgraph is.
    For example, there are two events with the same type of \textsc{Die} in the schema graph in Figure \ref{fig:example}, but the neighboring events of the first \textsc{Die} (e.g., \textsc{Detonate}, \textsc{Evacuate}, \textsc{Medical Intervention}) indicate that its subject is a victim of the IED bombing, while the neighboring events of the second \textsc{Die} (e.g., \textsc{Sentence}, \textsc{Indict}) indicate that it refers to the death of the attacker. 
    We apply a multi-layer graph neural network (GNN) to aggregate information from multi-hop neighboring nodes in the schema graph, and compute the correlation between the subgraph and the node in terms of their GNN representations.
    (2) \textit{Paths}.
    Note that modeling only neighboring node types is not able to identify the distance between the subgraph and the candidate node.
    It is also important to capture the set of paths connecting them, which can reveal the nature of their relation and enable the model to capture the distance between them.
    For example, in the schema graph in Figure \ref{fig:example}, it is clear that \textsc{Detonate} is more closely related to the first \textsc{Die} than the second \textsc{Die}, since \textsc{Detonate} and the first \textsc{Die} are directly connected, while the paths connecting \textsc{Detonate} and the second \textsc{Die} are much longer (e.g., \textsc{Detonate}-\textsc{Investigate}-\textsc{Identify}-\textsc{Indict}-\textsc{Sentence}-\textsc{Die}).
    We collect all paths connecting the candidate node and the subgraph, then use those paths to calculate their correlation.
    Finally, we combine these two modules together to predict the probability that the candidate node is missing for the subgraph.
    
    To solve the problem of lacking ground-truth missing events, we propose a self-supervised approach to construct training samples, which masks out a node from a given subgraph and treats the masked node as a missing event to be predicted.
    We also propose an inference algorithm specifically designed for event graph completion, which expands an incomplete instance graph over the schema by utilizing the structure of the schema graph.
    
    We conduct extensive experiments on four event graph datasets in the scenarios of IED bombing and disease outbreak.
    Experimental results demonstrate that our method achieves state-of-the-art performance in missing event prediction.
    For example, the F1 score of our method surpasses the best baseline method by $7.0\%$, $19.4\%$, $11.5\%$, and $4.3\%$ respectively, on the four datasets.
    Moreover, our ablation study and case study verify the effectiveness of the proposed neighbor module and path module.
    

    \begin{figure*}[t]
	   \centering
	   \includegraphics[width=0.99\textwidth]{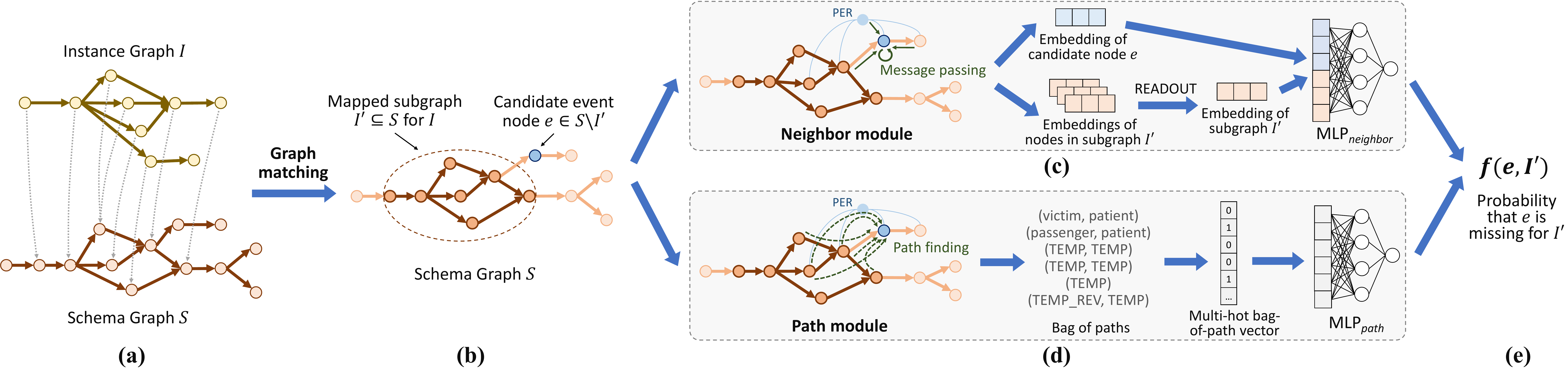}
	   \caption{Illustration of the model architecture. Entity nodes are omitted for the sake of clarity. (a) For an instance event graph $I$ and the event schema graph $S$, a two-stage subgraph matching algorithm is performed to map event nodes in $I$ to event nodes in $S$. (b) The instance graph $I$ is mapped to a subgraph $I' \subseteq S$. Our goal is to predict whether a candidate event node $e \in S \backslash I'$ is a missing node and thus should be added into $I'$. (c) The neighbor module. Message passing is performed on the schema graph to learn the representations of all nodes. An entity node \textsc{PER} (person) is drawn here to show that the message passing is performed on the entire schema graph. Embeddings of nodes in $I'$ are aggregated by a READOUT function to compute the embedding of $I'$. The embeddings of $e$ and $I'$ are fed into ${\rm MLP}_{neighbor}$. (d) The path module. All paths connecting $e$ and a node in $I'$ with length no more than a given threshold ($2$ in this example) are highlighted by green dotted arrows, which forms the bag-of-paths feature for the pair $(e, I')$. It is then transformed to a multi-hot bag-of-paths vector, which are fed into ${\rm MLP}_{path}$. (e) The outputs of ${\rm MLP}_{neighbor}$ and ${\rm MLP}_{path}$ are combined together to compute the final predicting probability $f(e, I')$.}
	   \label{fig:model}
    \end{figure*}

\section{Problem Formulation}

    We formulate the problem of schema-guided event graph completion as follows.
    Suppose we have a set of instance event graphs $\mathcal I = \{I_1, I_2, \cdots\}$, which are constructed from a set of articles with the same topic (e.g., car bombing).
    Each instance graph $I$ describes a complex event and consists of an event node set $\{e_i\}$ and an entity node set $\{v_i\}$, in which each node $e_i$ or $v_i$ is instantiated as a real event or entity.
    In addition, each event or entity is also associated with a specific event or entity type,
    and we use $\tau(\cdot)$ to denote the mapping function from a node to its type.
    Accordingly, there are three types of links in instance graphs:
    (1) event-event temporal link $\langle e_i, e_j \rangle$, which indicates that event $e_j$ happens chronologically after event $e_i$ (e.g., $\langle$\textsc{Detonate}, \textsc{Injure}$\rangle$);
    (2) event-entity link $\langle e_i, a, v_j \rangle$, which indicates that event $e_i$ has an argument role $a$, whose value is entity $v_j$ (e.g., $\langle$\textsc{Detonate}, \textsc{victim}, \textsc{Alice}$\rangle$);
    (3) entity-entity link $\langle v_i, r, v_j \rangle$, which indicates that there is a relation $r$ between entity $e_i$ and $e_j$ (e.g., $\langle$\textsc{John}, \textsc{located at}, \textsc{The Square}$\rangle$).
    We also use function $\tau(\cdot)$ to denote the type of a link.
    Specifically, $\tau(\langle e_i, e_j \rangle) =$ \textsc{TEMP},\footnote{We use \textsc{TEMP} to denote the type of a temporal link, and \textsc{TEMP\_REV} to denote the type of a reversed temporal link.} $\tau(\langle e_i, a, v_j \rangle) = a$, and $\tau(\langle v_i, r, v_j \rangle) = r$.
    
    Moreover, we also have a schema graph $S$ available, which is a generic representation of instance graphs in $\mathcal I$ and characterizes their typical structure.
    The format of the schema graph $S$ is similar to instance graphs, while the only difference is that nodes in $S$ are not instantiated but are only specified by event or entity types.
    
    Given the training instance graphs $\mathcal I$ and the schema graph $S$, we aim to learn a model that is able to predict missing events for a new and incomplete instance graph.
    Specifically, for a new instance graph $I$ and a candidate event node $e \in S$, our model aims to predict the probability that $e$ is a missing event for $I$.\footnote{It is important to predict not only the type of missing events but also their arguments as well as temporal connections to existing event nodes. We will discuss the details of argument/temporal link completion in Section \ref{sec:inference}.} 

\section{Our Approach}
    Given an incomplete instance event graph as well as the event schema graph, we first perform subgraph matching between them to map the instance graph to a subgraph of the schema graph (Section \ref{sec:graph_matching}).
    Then we design a neighbor module (Section \ref{sec:neighbors}) and a path module (Section \ref{sec:paths}) to compute the correlation between the mapped subgraph and a candidate event node in the schema.
    We also discuss the training (Section \ref{sec:training}) and inference (Section \ref{sec:inference}) procedures of the proposed model.
    The architecture of our proposed model is shown in Figure \ref{fig:model}.

    \subsection{Subgraph Matching}
    \label{sec:graph_matching}
        Since our goal is to predict whether a candidate event node $e$ from the schema graph $S$ is a missing node for an instance graph $I$, the first step is therefore to perform subgraph matching between $S$ and $I$.
        In this way, $I$ can be mapped to a subgraph of $S$, which enables us to compute the correlation between the instance graph $I$ and the candidate event node $e$ within the scope of schema graph $S$.
        
        \xhdr{Integer Programming}
        We can use a binary assignment matrix ${\bf X} \in \{0, 1\}^{|I| \times |S|}$ to represent the matching solution, in which ${\bf X}_{ij} = 1$ if and only if node $i \in I$ matches node $j \in S$.
        We also use $s_N(i, j)$ to denote the pairwise node similarity for $i \in I$ and $j \in S$, and $s_L(\langle i, j \rangle, \langle k, l \rangle)$ to denote the pairwise link similarity function for $\langle i, j \rangle \in I \times I$ and $\langle k, l \rangle \in S \times S$.
        Then the subgraph matching problem can be formulated as finding the assignment matrix that maximizes the total matching score:
        \begin{equation}
        \label{eq:ip}
        \begin{split}
            &\max_{\bf X} \sum_{{\bf X}_{ij}=1} s_N(i, j) + \sum_{\substack{{\bf X}_{ik}=1 \\ {\bf X}_{jl}=1}} s_L(\langle i, j \rangle, \langle k, l \rangle) \\
            &s.t.
            \begin{cases}
                \ {\bf X} \in \{0, 1\}^{|I| \times |S|} \\
                \ \sum_{j=1}^{|S|} {\bf X}_{ij} \leq 1, \ \text{for} \ i = 1, \cdots, |I|,
            \end{cases}
        \end{split}
        \end{equation}
        where the constraints define a many-to-one node mapping: a node in $I$ can be mapped to at most one node in $S$, while a node in $S$ can be mapped to multiple nodes in $I$.
        This is because an instance graph may have repeated event types (e.g., two \textsc{Injure} events in Figure \ref{fig:example}), which should be mapped to the same event node in the schema.
    
        It is worth noting that Eq. (\ref{eq:ip}) defines a 0-1 integer programming problem \citep{cho2014finding}, which is NP-complete.
        To improve the time efficiency of subgraph matching, we notice the following two facts:
        (1) The instance graph and the schema graph are both event-node-centric.
        Therefore, we can focus on matching event nodes in the two graphs, which will greatly reduce the entire searching space.
        Once event nodes are matched, entity nodes can be naturally matched based on their event argument roles.
        (2) Event types play a decisive role when determining whether two event nodes are matched.
        For example, in Figure \ref{fig:example}, the \textsc{die:0} node in the instance graph should be matched to \textsc{Die} in the schema graph rather than \textsc{Injure} since they have the same event type, even if the arguments of \textsc{die:0} are different from \textsc{Die} but are exactly the same as \textsc{Injure}.
        
        \xhdr{A Two-Stage Heuristic Subgraph Matching Scheme}
        Taking into consideration the above two facts, we propose a two-stage heuristic subgraph matching scheme between instance and schema graphs.
        For an event node $e_i \in I$ to be matched, we first identify the event node(s) in $S$ whose type is the same as $e_i$:\footnote{If event types in instance graphs and the schema do not follow the same naming style and thus cannot be compared directly, we can use embedding-based fuzzy matching: $e^* = \arg \max_{e_j \in S} \text{LM} \big( \tau(e_i) \big)^\top \text{LM} \big( \tau(e_j) \big)$, where $\text {LM}(\cdot)$ is a language model encoder such as BERT \citep{devlin2018bert}.}
        \begin{equation}
        \label{eq:step1}
            E_i = \{e_j \in S \ | \ \tau(e_j) = \tau(e_i) \},
        \end{equation}
        Based on the size of $E_i$, we have the following three cases:
        \begin{itemize}
            \item $|E_i| = 0$. This means that $e_i$ will not be matched to any event node in $S$ (e.g., the \textsc{contact:0} node in Figure \ref{fig:example}).
            \item $|E_i| = 1$. In this case, we end up with a unique node in $S$, which is taken as the matching node for $e_i$ (e.g., the \textsc{assemble:0} node in Figure \ref{fig:example}).
            \item $|E_i| > 1$, i.e., there are more than one event node in $S$ that can be matched to $e_i$, since there may be multiple events in $S$ that have the same event type. For example, the \textsc{die:0} node in Figure \ref{fig:example} can be matched to two \textsc{Die} nodes.
        \end{itemize}
        
        For the last case above, we design an additional stage for event matching, which is based on the similarity of neighbor node types.
        Specifically, for each event $e_j \in E_i$, we identify the set of types of $e_j$'s one-hop previous events $\mathcal P$, one-hop following events $\mathcal F$, and argument roles $\mathcal A$, in schema graph $S$:
        \begin{equation}
        \label{eq:type_set}
        \begin{split}
            \mathcal P(e_j; S) &= \{\tau(e_k) \ | \ \langle e_k, e_j \rangle \in S\},\\
            \mathcal F(e_j; S) &= \{\tau(e_k) \ | \ \langle e_j, e_k \rangle \in S\},\\
            \mathcal A(e_j; S) &= \{a \ | \ \langle e_j, a, v_k \rangle \in S\}.
        \end{split}
        \end{equation}
        
        We can also identify the above three sets for $e_i$ in instance graph $I$, i.e., $\mathcal P(e_i; I)$, $\mathcal F(e_i; I)$, and $\mathcal A(e_i; I)$.
        Then we calculate the Jaccard index\footnote{The Jaccard index between two sets is defined as $J(A, B) = \frac{|A \cap B|}{|A \cup B|}$. If $A$ and $B$ are both empty, $J(A, B) = 1$.} $J$ of the three corresponding set pairs, and select the node with the highest total Jaccard index as the matching result:
        \begin{equation}
        \label{eq:step2}
        \begin{split}
            e^* = \arg \max_{e_j \in E_i} \ &J \big( \mathcal P(e_j; S), \mathcal P(e_i; I) \big)
            \ + \ J \big( \mathcal F(e_j; S), \mathcal F(e_i; I) \big) \\
            + \ & J \big( \mathcal A(e_j; S), \mathcal A(e_i; I) \big).
        \end{split}
        \end{equation}

        Note that in this two-stage matching scheme, the first stage (Eq. (\ref{eq:step1})) and the second stage (Eq. (\ref{eq:step2})) exactly correspond to the two terms in the objective function in Eq. (\ref{eq:ip}), i.e., node similarity and link similarity.
        In most cases, this two-step matching scheme will return a unique matching node $e^* \in S$ for the input node $e_i \in I$.
        But if there are still more than one nodes in $S$ that have the highest total Jaccard similarity with $e_i$, we will randomly select one as the matching result.
        
        Here we analyze the time complexity of our proposed subgraph matching algorithm.
        We use $N_I$ and $N_S$ to denote the number of event nodes in $I$ and $S$, respectively, and $d$ to denote their average node degree. 
        For a given event $e_i \in I$, the complexity of calculating $E_i$ in Eq. (\ref{eq:step1}) is $O(N_S)$, and the complexity of calculating $\mathcal P(e_i; I)$, $\mathcal F(e_i; I)$, and $\mathcal A(e_i; I)$ is $d$.
        In the worst case where $|E_i| = N_S$, we need to calculate Eq. (\ref{eq:type_set}) as well as the Jaccard index for all nodes in $S$, whose complexity is $O(N_S (d + d)) = O(N_S d)$.
        Therefore, the total complexity is $O(N_I (N_S + d) N_S d) = O (N_I N_S^2 d)$, which is polynomial w.r.t. the size of $I$ and $S$.

    \subsection{Neighbors}
    \label{sec:neighbors}
        After subgraph matching between the instance graph $I$ and the schema graph $S$, $I$ is mapped to $I'$, which is a subset of event nodes in $S$.
        Our goal is therefore to learn a predicting function $f(e, I')$, which outputs the probability of whether a new event $e \in S \backslash I'$ is a missing node for $I' \subseteq S$.
        In this subsection, we measure the correlation between $I'$ and $e$ in terms of their neighbors.
        
        To learn the representation of nodes as well as subgraphs in the schema, we choose GNNs, which utilize graph structure and node features to learn a representation vector for each node, as our base model.
        Typical GNNs follow a neighborhood aggregation strategy, which iteratively updates the representation of an atom by aggregating representations of its neighbors.
        Formally, the $k$-th layer of a GNN is:
        \begin{equation}
        \label{eq:gnn}
            {\bf h}_i^k = {\rm AGG} \left( \big \{ {\bf h}_j^{k-1} \big \}_{j \in \mathcal N(i) \cup \{i\}} \right),
        \end{equation}
        where ${\bf h}_i^k$ is the representation vector of node $i \in S$ at the $k$-th layer (${\bf h}_i^0$ is initialized as the one-hot vector of $i$'s node type), and $\mathcal N(i)$ is the set of nodes directly connected to $i$.
        The choice of AGG function is essential to designing GNNs, and a number of GNN architectures have been proposed \citep{kipf2017semi,hamilton2017inductive,velivckovic2018graph}.
        For example, in Graph Convolutional Networks (GCN) \citep{kipf2017semi}, AGG is implemented as weighted average with respect to the inverse of square root of node degrees:
        \begin{equation}
            {\bf h}_i^k = \sigma \Big( {\bf W}^k \sum\nolimits_{j \in \mathcal N(i) \cup \{i\}} \alpha_{ij} {\bf h}_j^{k-1} + {\bf b}^k \Big),
        \end{equation}
        where $\alpha_{ij} = 1 / \sqrt{|\mathcal N(i)| \cdot |\mathcal N(j)|}$, ${\bf W}^k$ and ${\bf b}^k$ are a learnable matrix and bias, respectively, and $\sigma$ is an activation function.
        We use GCN as the implementation of GNN, while the performance of other GNN models is reported in Appendix \ref{app:hs}.
        
        Suppose the number of GNN layers is $K$.
        The final embeddings of event $e$ and events in the subgraph $I'$ are therefore ${\bf h}_e^K$ and $\big \{ {\bf h}_{e_i}^K \big \}_{e_i \in I'}$, respectively.
        Then, a readout function is used to aggregate event embeddings in $I'$ and output the embedding of $I'$:
        \begin{equation}
            {\bf h}_{I'}^K = {\rm READOUT} \left( \big \{ {\bf h}_{e_i}^K \big \}_{e_i \in I'} \right).
        \end{equation}
        The READOUT function can be summation, average, or a more sophisticated attention-based aggregation, since the importance of events in a subgraph may be different with respect to the given event $e$:
        \begin{equation}
        \label{eq:att}
            {\bf h}_{I'}^K = \sum\nolimits_{e_i \in I'} \beta_i {\bf h}_{e_i}^K,
        \end{equation}
        where $\beta_i = \frac{{{\bf h}_{e_i}^K}^\top {\bf h}_e^K}{\sum_{e_i \in I'} {{\bf h}_{e_i}^K}^\top {\bf h}_e^K}$ is the attention weight.
        We will report the performance of different READOUT functions in Appendix \ref{app:hs}.
        
        Finally, the embeddings of $e$ and $I'$ are concatenated, followed by a Multi-Layer Perceptron (MLP) to predict the probability that $e$ is missing for $I'$:
        \begin{equation}
        \label{eq:neighbor}
            p_{neighbor} (e, I') = {\rm MLP}_{neighbor} \left( \big [{\bf h}_e^K, {\bf h}_{I'}^K \big] \right).
        \end{equation}

    \subsection{Paths}
    \label{sec:paths}
        Note that when using GNN to process node neighbors for the mapped subgraph $I'$ and the candidate event node $e$, we use node types as the initial node feature, which leads to a potential issue that our model is not able to identify the distance between $I'$ and $e$.
        For example, think of an extreme case where we have two event nodes $e_1$ and $e_2$ in the schema graph, whose neighbor structures within $K$ hops are exactly the same 
        in terms of node types.
        In this case, the final embeddings of event $e_1$ and $e_2$ are also the same if we use a GNN with no more than $K$ layers to process their neighbors.
        This implies that, for any subgraph $S' \subseteq S$, the predicted probability of $p_{neighbor} (e_1, S')$ and $p_{neighbor} (e_2, S')$ will also be the same, according to Eq. (\ref{eq:neighbor}).
        However, this is not always reasonable, since a subgraph $S'$ may be much closer to $e_1$ than $e_2$ in the schema graph, thus $p_{neighbor} (e_1, S')$ should be much larger than $p_{neighbor} (e_2, S')$.
        
        To make our model capture the distance information, we model the connectivity pattern between a subgraph and a node, which is characterized by paths connecting them in the schema graph.
        Specifically, a path connecting two nodes $s$ and $t$ is a sequence of nodes and edges:
        \begin{equation}
            s \xrightarrow{\langle s, i \rangle} i \xrightarrow{\langle i, j \rangle} j \cdots k \xrightarrow{\langle k, t \rangle} t,
        \end{equation}
        where $\langle i, j \rangle$ is the edge connecting node $i$ and $j$, and each node in the path is unique\footnote{The nodes in a path are required to be unique because a loop does not provide any additional information and thus should be cut off from the path.}.
        In this work, we use the types of edges in a path to represent the path, i.e.,
        \begin{equation}
            \big( \tau(\langle s, i \rangle), \tau(\langle i, j \rangle), \cdots, \tau(\langle k, t \rangle) \big).
        \end{equation}
        We use $\mathcal P_{s \rightarrow t}^{\leq L}$ to denote the set of all paths connecting $s$ and $t$ with length of no more than $L$, where $L$ is a given hyperparameter.
        For example, as shown in Figure \ref{fig:example}, there are two paths with length of one or two that connect \textsc{Transport} and \textsc{Detonate} in the schema graph, i.e., $\mathcal P_{\textsc{Transport} \rightarrow \textsc{Detonate}}^{\leq 2} = \{ (\textsc{TEMP}), (\textsc{destination}, \textsc{place}) \}$.
        
        For an given event $e$ and a subgraph $I'$ in the schema graph, we collect all paths connecting $e$ and every event node in $I'$ as the path set for $(e, I')$:
        \begin{equation}
            \mathcal P_{e \rightarrow I'}^{\leq L} = \bigcup_{e_i \in I'} \mathcal P_{e \rightarrow e_i}^{\leq L},
        \end{equation}
        which is then transformed into a multi-hot bag-of-paths vector ${\bf p}_{e \rightarrow I'}^{\leq L}$, where each entry indicates that if a particular path exists in $\mathcal P_{e \rightarrow I'}^{\leq L}$.
        We use another MLP to take ${\bf p}_{e \rightarrow I'}^{\leq L}$ as input and output the probability that $e$ is missing for $I'$:
        \begin{equation}
        \label{eq:path}
            p_{path} (e, I') = {\rm MLP}_{path} \left( {\bf p}_{e \rightarrow I'}^{\leq L} \right).
        \end{equation}
        
        Finally, our predicting function $f(e, I')$ is implemented by combining the output of the neighbor module in Eq. (\ref{eq:neighbor}) and the path module in Eq. (\ref{eq:path}):
        \begin{equation}
            f(e, I') = \frac{1}{2} \left( p_{neighbor} (e, I') + p_{path} (e, I') \right).
        \end{equation}

    \subsection{Training}
    \label{sec:training}
        A potential issue of training the proposed model is the lack of ground-truth for predicting missing events for a given instance graph.
        Therefore, we propose a \textit{self-supervised} loss as the training target.
        Specifically, we first map each instance graph $I$ in the training data to the schema graph $S$ and get the matched subgraph $I'$.
        Then for each event node $e \in I'$, we mask $e$ 
        out from $I'$ and try to predict $e$ using the rest of $I'$.
        In other words, we treat $(e, I'\backslash e)$ as a positive training sample for each $e \in I'$.
        Meanwhile, we can randomly sample an event node outside $I'$, i.e., $e \in S\backslash I'$, and treat $(e, I')$ as a negative sample.
        The total loss function is therefore as follows:
        \begin{equation}
        \label{eq:loss}
            L = \frac{1}{|\mathcal I|} \sum_{I \in \mathcal I} \frac{1}{|S|} \Bigg( \sum_{e \in I'} \mathcal C \Big( f(e, I' \backslash e), 1 \Big) + \sum_{e \in S \backslash I'} \mathcal C \Big( f(e, I'), 0 \Big) \Bigg),
        \end{equation}
        where $\mathcal I$ is the set of training instance graphs, $I'$ is a subgraph of $S$ that $I$ is mapped to, and $\mathcal C$ is cross-entropy loss.
        In Eq. (\ref{eq:loss}), the first term is the loss for positive samples, and the second term is the loss for negative samples.
        
        Note that the training data constructed by Eq. (\ref{eq:loss}) may be unbalanced, i.e., the total numbers of positive and negative samples may be significantly different.
        In this case, we can use downsampling strategy to re-balance the dataset.

    \subsection{Inference}
    \label{sec:inference}
        The goal of inference is to complete an input instance event graph, i.e., predict all the missing nodes.
        This is similar to \textit{set expansion} \cite{shen2017setexpan,zaheer2017deep,huang2020guiding}, which aims to expand a small set of seed elements into a complete set of elements that belong to the same semantic class.
        But note that there exists an intrinsic graph structure among elements in our problem.
        Therefore, we propose an inference algorithm specifically for schema-guided event graph completion.

        \begin{algorithm}[t]
            \caption{Inference procedure}
            \small
	        \label{alg:training}
	        \KwIn{An incomplete instance graph $I$, the schema graph $S$, the trained model $f(e, I'): S \times 2^S \mapsto [0, 1]$}
	        \KwOut{The completed graph $\hat I$}
	        
	        $\hat I \leftarrow I$\;
	        Map $I$ to $I' \subseteq S$ using the two-stage subgraph matching algorithm presented in Section \ref{sec:graph_matching}\;
	        Candidate event node set $C \leftarrow S \backslash I'$\;
	        \While{$C$ is not empty}{
    	        $c' \leftarrow \arg \max_{e \in C: \ dist(e, I')=1} f(e, I')$\;
    	        \If{$f(c', I') > Threshold$}{
    	            $I' \leftarrow I' \cup \{c'\}$, $C \leftarrow C \backslash c'$\;
    	            Add a new event node $c$ to $\hat I$ whose type is the same as $c'$\;
    	            \For{$n'$ in $c'$'s neighbor events and neighbor entities in $S$}{
    	                \If{$n'$ has a matching node $n$ in $\hat I$}{
    	                    Add a temporal or argument link between $c$ and $n$ in $\hat I$ according to the link between $c'$ and $n'$ in $S$\;
    	                }
    	            }
    	        }
    	        \Else{
                \textbf{break}}
            }
	        \Return{$\hat I$}
        \end{algorithm}

        \begin{table*}[t]
			\centering
			\small
			\setlength{\tabcolsep}{10pt}
			\begin{tabular}{c||c|c|c|c}
				\hline
				\hline
				Dataset & Car-Bombings & IED-Bombings & Suicide-IED & Pandemic \\
				\hline
				\# train/validation/test instance graphs & 75 / 9 / 10 & 88 / 11 / 12 & 176 / 22 / 22 & 40 / 5 / 6 \\
				\# train/validation/test samples & 2,368 / 288 / 320 & 2,904 / 363 / 396 & 5,808 / 726 / 726 & 3,200 / 400 / 480 \\
				\hline
				Corresponding schema name & Car-IED & \multicolumn{2}{c|}{General-IED} & Disease-Outbreak \\
				\# event/entity nodes & 32 / 134 & \multicolumn{2}{c|}{33 / 140} & 102 / 17 \\
				\# event-event/event-entity/entity-entity links & 41 / 138 / 261 & \multicolumn{2}{c|}{42 / 143 / 530} & 200 / 75 / 1 \\
				\hline
				\hline
			\end{tabular}
			\vspace{0.05in}
			\caption{Statistics of the four datasets and three event schemas.}
			\vspace{-0.1in}
			\label{table:dataset}
		\end{table*}
        
        The inference procedure is presented in Algorithm \ref{alg:training}, which takes as input an incomplete instance graph $I$, the schema gaph $S$, and the trained model $f(e, I')$, then output the completed graph $\hat I$.
        The first step is to use the subgraph matching algorithm introduced in Section \ref{sec:graph_matching} to map $I$ to a subgraph $I' \subseteq S$ (line 2), and the candidate event node set $C$ is initialized as the set of nodes outside subgraph $I'$ (line 3).
        Then we select a node from the candidate set $C$ as the predicted missing event repeatedly (lines 4-13).
        At each iteration, we first identify the subset of the candidate set where nodes are neighbors of $I'$ (i.e., $e \in C$ and $dist(e, I') = 1$), then select a node $c'$ from this subset whose probability $f(c', I')$ is the highest, as the candidate event node at this iteration (line 5).
        The reason of restricting $c'$ to be the neighbor of $I'$ is that, $c'$ can thus be linked directly to $I'$ through temporal links, so that the instance graph after completion will still be connected.
        Based on the value of $f(c', I')$, we have the following two cases:
		
		\begin{itemize}
		    \item If $f(c', I')$ is greater than a given threshold (line 6), such as $0.5$, $c'$ is taken as the predicted missing event at this iteration. We then update $I'$ and $C$ by adding $c'$ to $I'$ and removing $c'$ from $C$, respectively (line 7), and update $\hat I$ by adding the missing event and its associated links to $\hat I$ (lines 8-11). Specifically, we add a new event node $c$ to $\hat I$ whose type is the same as $c'$ (line 8), then for each event and entity node $n' \in S$ that is connected to $c'$, if $n'$ has a matching node $n \in I$, we add a temporal link or argument link between $c$ and $n$ according to the link between $c'$ and $n'$ in $S$ (lines 9-11). In other words, we ``copy'' the links associated to the predicted event from the schema to the instance graph.
		
            \item Otherwise (line 12), the scores of all nodes in the candidate set do not exceed the threshold (note that $f(c', I')$ is the largest among $C$). We can then terminate the inference procedure (line 13).
        \end{itemize}
		
		The inference loop is repeated until the candidate set $C$ is empty or the probability $f(c', I')$ in the current loop is no larger than the threshold.
        Finally, $\hat I$ is returned as the completed event graph for $I$.
        
        It is worth noting that missing events are predicted according to the ascending order of their distance to the subgraph $I'$, which ensures that the original instance graph $I$ is expanded over the schema in a natural, from-the-inside-out manner.
        In addition, note that the predicted missing event at one iteration is immediately added into $I'$, which is then used to compute the probability $f$ for the next iteration.
        Such a \textit{bootstrapping} strategy allows the model to track the up-to-date instance graph and predict missing event nodes as many as possible.

\section{Experiments}
    In this section, we evaluate the proposed model, and present its performance on four datasets.
    The code and datasets are provided in the supplementary material.
    
    \subsection{Datasets}
        We conduct experiments on four event instance graph datasets: \textit{Car-Bombing}s, \textit{IED-Bombings}, \textit{Suicide-IED}, and \textit{Pandemic}.
        The first three datasets are constructed by \citet{li2021future}, which consist of complex events related to IED bombing.
        The last dataset Pandemic is constructed by us.
        Specifically, we use RESIN \cite{wen2021resin}, a cross-document information extraction and event tracking system, to process news articles mentioned in the references of Wikipedia articles related to pandemic, e.g., 2002-2004 SARS outbreak, then construct an instance event graph for each disease outbreak.
        
        In addition, we use three complex event schemas for the four datasets.
        The first schema \textit{Car-IED} describes the scenario of bombing caused by car IEDs that is detonated in an automobile or other vehicles.
        We use Car-IED schema for Car-Bombings dataset.
        The second schema \textit{General-IED} describes the scenario of general IED bombing, which is used for IED-Bombings and Suicide-IED datasets.
        The last schema is \textit{Disease-Outbreak}, which describes the spread of pandemics in a given population as well as the response of the authority, the public, etc.
        We use Disease-Outbreak schema for Pandemic dataset.
        The first two schemas are developed by \citet{li2021future}, while the last schema is manually curated by us.
        The statistics of the four datasets and the three schemas are presented in Table \ref{table:dataset}.

    \subsection{Baseline Methods}
        We compare our method with the following baseline methods:
        
        \begin{itemize}
            \item \textit{AddAll}, which treats all events that exist in the schema but not exist in an instance graph as missing events. In other words, it treats all test samples as positive.
            \item \textit{AddNeighbor}, which treats all events in the schema graph that are adjacent to the mapped subgraph of an instance graph as missing events. In other words, it treats a test sample $(e, I')$ as positive if and only if $e$ is the neighbor of at least one event in $I'$.
            \item \textit{ID-MLP}, which concatenates the one-hot ID vector of $e$ and the multi-hot ID vector of $I'$ for the pair $(e, I')$ as input, followed by an MLP to predict the probability that $e$ is a missing event for $I'$.
            \item \textit{Type-MLP}, which is similar to ID-MLP, but the input is the concatenated vector of the one-hot event type vector of $e$ and the multi-hot event type vector of $I'$ for the pair $(e, I')$.
            \item \textit{TransE} \cite{bordes2013translating} is a classic knowledge graph completion method, which assumes that ${\bf h} + {\bf r} \approx {\bf t}$ for each triplet $(h, r, t)$ in the knowledge graph, where $h$ and $t$ are the head and tail entities, respectively, and $r$ is the relation.
            \item \textit{RotatE} \cite{sun2018rotate} is another state-of-the-art knowledge graph completion method similar to TransE, but it models entity and relation embeddings in the complex number space.
        \end{itemize}
        
        The implementation details of baselines are presented in Appendix \ref{app:implementation}.
        In addition, we also conduct extensive ablation study and propose two reduced versions of our model, \textit{\alg-Neighbor} and \textit{\alg-Path}, which only use neighbor information and path information, respectively, to test the performance of the two components separately.

    \subsection{Experimental Setup}
        We evaluate our method on two tasks: \textit{binary classification} and \textit{graph completion}.
        
        For the binary classification task, given an instance graph $I$ in the test set, we first map $I$ to the corresponding schema graph $S$ and get the mapped subgraph $I' \subseteq S$.
        Then for each event $e \in I'$, we treat $(e, I'\backslash e)$ as a positive test sample, and for each event $e \in S\backslash I'$, we treat $(e, I')$ as a negative test sample.
        We use \textit{Accuracy} and \textit{AUC} as the evaluation metrics.

        \begin{table*}[t]
			\centering
			\small
			\setlength{\tabcolsep}{4pt}
			\begin{tabular}{c|cc|cc|cc|cc}
				\hline
				\hline
				Dataset & \multicolumn{2}{c|}{Car-Bombings} & \multicolumn{2}{c|}{IED-Bombings} & \multicolumn{2}{c|}{Suicide-IED} & \multicolumn{2}{c}{Pandemic} \\
				\hline
				Metrics & Accuracy & AUC & Accuracy & AUC & Accuracy & AUC & Accuracy & AUC \\
				\hline
				AddAll & 0.553 & 0.500 & 0.353 & 0.500 & 0.404 & 0.500 & 0.190 & 0.500 \\
				AddNeighbor & 0.594 & 0.569 & 0.586 & 0.641 & 0.609 & 0.645 & 0.648 & 0.568 \\
				ID-MLP & 0.784 $\pm$ 0.032 & 0.889 $\pm$ 0.018 & 0.721 $\pm$ 0.008 & 0.857 $\pm$ 0.005 & 0.792 $\pm$ 0.007 & 0.883 $\pm$ 0.006 & 0.912 $\pm$ 0.006 & 0.963 $\pm$ 0.006 \\
				Type-MLP & 0.802 $\pm$ 0.012 & 0.898 $\pm$ 0.008 & 0.724 $\pm$ 0.021 & 0.858 $\pm$ 0.009 & 0.799 $\pm$ 0.007 & 0.890 $\pm$ 0.004 & 0.914 $\pm$ 0.007 & 0.964 $\pm$ 0.004 \\
				TransE & 0.795 $\pm$ 0.020 & 0.890 $\pm$ 0.019 & 0.733 $\pm$ 0.038 & 0.845 $\pm$ 0.031 & 0.784 $\pm$ 0.014 & 0.876 $\pm$ 0.009 & 0.905 $\pm$ 0.012 & 0.952 $\pm$ 0.010 \\
				RotatE & 0.743 $\pm$ 0.040 & 0.860 $\pm$ 0.045 & 0.708 $\pm$ 0.026 & 0.865 $\pm$ 0.006 & 0.734 $\pm$ 0.013 & 0.851 $\pm$ 0.011 & 0.882 $\pm$ 0.015 & 0.940 $\pm$ 0.011 \\
				\hline
				\alg & 0.828 $\pm$ 0.017 & \textbf{0.923} $\pm$ 0.003 & \textbf{0.815} $\pm$ 0.012 & \textbf{0.889} $\pm$ 0.002 & \textbf{0.824} $\pm$ 0.003 & \textbf{0.900} $\pm$ 0.005 & \textbf{0.928} $\pm$ 0.005 & \textbf{0.976} $\pm$ 0.001 \\
				\alg-Neighbor & \textbf{0.836} $\pm$ 0.014 & 0.921 $\pm$ 0.004 & 0.800 $\pm$ 0.021 & 0.888 $\pm$ 0.006 & 0.818 $\pm$ 0.006 & 0.898 $\pm$ 0.003 & 0.910 $\pm$ 0.003 & 0.956 $\pm$ 0.005 \\
				\alg-Path & 0.809 $\pm$ 0.008 & 0.893 $\pm$ 0.007 & 0.795 $\pm$ 0.009 & 0.861 $\pm$ 0.004 & 0.818 $\pm$ 0.011 & 0.883 $\pm$ 0.008 & 0.919 $\pm$ 0.006 & 0.966 $\pm$ 0.002 \\
				\hline
				\hline
			\end{tabular}
			\vspace{0.05in}
			\caption{Mean and standard deviation of all methods on binary classification task. The best results are highlighted in bold.}
			\vspace{-0.15in}
			\label{table:result_1}
		\end{table*}
		
		\begin{table*}[t]
			\centering
			\small
			\setlength{\tabcolsep}{4pt}
			\begin{tabular}{c|cc|cc|cc|cc}
				\hline
				\hline
				Dataset & \multicolumn{2}{c|}{Car-Bombings} & \multicolumn{2}{c|}{IED-Bombings} & \multicolumn{2}{c|}{Suicide-IED} & \multicolumn{2}{c}{Pandemic} \\
				\hline
				Metrics & Jaccard Index & F1 & Jaccard Index & F1 & Jaccard Index & F1 & Jaccard Index & F1 \\
				\hline
				AddAll & 0.173 & 0.278 & 0.083 & 0.150 & 0.099 & 0.173 & 0.032 & 0.061 \\
				AddNeighbor & 0.165 $\pm$ 0.009 & 0.255 $\pm$ 0.019 & 0.109 $\pm$ 0.010 & 0.187 $\pm$ 0.016 & 0.124 $\pm$ 0.006 & 0.210 $\pm$ 0.011 & 0.047 $\pm$ 0.011 & 0.087 $\pm$ 0.020 \\
				ID-MLP & 0.332 $\pm$ 0.019 & 0.466 $\pm$ 0.022 & 0.178 $\pm$ 0.033 & 0.282 $\pm$ 0.036 & 0.282 $\pm$ 0.024 & 0.391 $\pm$ 0.028 & 0.292 $\pm$ 0.050 & 0.380 $\pm$ 0.050 \\
				Type-MLP & 0.387 $\pm$ 0.018 & 0.522 $\pm$ 0.022 & 0.187 $\pm$ 0.060 & 0.289 $\pm$ 0.075 & 0.336 $\pm$ 0.059 & 0.428 $\pm$ 0.053 & 0.303 $\pm$ 0.070 & 0.412 $\pm$ 0.064 \\
				TransE & 0.349 $\pm$ 0.022 & 0.467 $\pm$ 0.029 & 0.196 $\pm$ 0.056 & 0.286 $\pm$ 0.063 & 0.302 $\pm$ 0.068 & 0.401 $\pm$ 0.064 & 0.273 $\pm$ 0.066 & 0.337 $\pm$ 0.082 \\
				RotatE & 0.291 $\pm$ 0.065 & 0.418 $\pm$ 0.066 & 0.146 $\pm$ 0.037 & 0.235 $\pm$ 0.033 & 0.208 $\pm$ 0.024 & 0.301 $\pm$ 0.031 & 0.266 $\pm$ 0.058 & 0.319 $\pm$ 0.062 \\
				\hline
				\alg & \textbf{0.458} $\pm$ 0.050 & \textbf{0.592} $\pm$ 0.054 & \textbf{0.348} $\pm$ 0.031 & \textbf{0.483} $\pm$ 0.032 & \textbf{0.449} $\pm$ 0.038 & 0.543 $\pm$ 0.050 & \textbf{0.331} $\pm$ 0.043 & \textbf{0.455} $\pm$ 0.073 \\
				\alg-Neighbor & 0.420 $\pm$ 0.052 & 0.564 $\pm$ 0.054 & 0.328 $\pm$ 0.091 & 0.434 $\pm$ 0.103 & 0.444 $\pm$ 0.074 & \textbf{0.555} $\pm$ 0.061 & 0.256 $\pm$ 0.009 & 0.389 $\pm$ 0.007 \\
				\alg-Path & 0.416 $\pm$ 0.072 & 0.538 $\pm$ 0.067 & 0.288 $\pm$ 0.055 & 0.424 $\pm$ 0.071 & 0.378 $\pm$ 0.056 & 0.498 $\pm$ 0.055 & 0.311 $\pm$ 0.075 & 0.430 $\pm$ 0.078 \\
				\hline
				\hline
			\end{tabular}
			\vspace{0.05in}
			\caption{Mean and standard deviation of all methods on graph completion task. The best results are highlighted in bold.}
			\vspace{-0.15in}
			\label{table:result_2}
		\end{table*}

		For the graph completion task, given an instance graph $I$ in the test set, we first randomly hide $10\%$ of event nodes from $I$ and treat the masked nodes as the ground-truth set, then run Algorithm \ref{alg:training} (the threshold in line 6 is set to $0.5$) to complete the graph.
        Specifically, we predict the set of missing events using the remaining $90\%$ of event nodes in the graph, then compare the predicted set and the ground-truth set by calculating their \textit{Jaccard Index} and $F1$ score.
        
        We report the performance of our model on the test set when AUC on the validation set is maximized.
        Each experiment is repeated five times, and we report the mean and standard deviation of the results.
		The implementation details and hyperparameter settings are presented in Appendix \ref{app:implementation}, and the sensitivity of our model to hyperparameter settings is  presented in Appendix \ref{app:hs}.
    
    \subsection{Results}
        \xhdr{Comparison with Baselines}
            The results of binary classification task and graph completion task are presented in Table \ref{table:result_1} and Table \ref{table:result_2}, respectively.
            Our method achieves the best performance on all datasets.
            Specifically, the AUC of our method \alg in binary classification task surpasses the best baseline by $2.5\%$, $2.4\%$, $1.0\%$, and $1.2\%$, respectively, on the four datasets, and the F1 of \alg in graph completion task surpasses the best baseline by $7.0\%$, $19.4\%$, $11.5\%$, and $4.3\%$, respectively, on the four datasets (all the numbers are absolute gains).
            We notice that the advantages of our method are much more significant in graph completion task, which demonstrate the superiority of our method in the real graph completion scenario.
            
            We also observe that, in most cases, the two reduced versions of method \alg-Neighbor and \alg-Path already perform quite well and beat all the baseline methods.
            Combining them together usually leads to even better performance.

        \xhdr{Impact of Schema Quality}
            Since our proposed method is based on event schemas, we study how the quality of schemas will impact the model performance.
            We randomly perturb the original Car-IED schema for Car-Bombings dataset, and present the model performance in Figure \ref{fig:schema_quality}.
            The performance of the path module (\alg-path) is basically unchanged when the percentage of changed edges is small ($\leq 20\%$), but drops significantly when more edges in the schema are changed ($\geq 30\%$).
            In contrast, the neighbor module (\alg-neighbor) is more resistant to the noise in schema graph.
            As a result, the whole \alg model is able to basically maintain its performance even when half of the edges in the schema graph are changed.

    \subsection{Case Study}
    \label{sec:cs}
\begin{figure*}[t]
		\centering
		\begin{minipage}[t]{0.29\textwidth}
            \includegraphics[width=\textwidth]{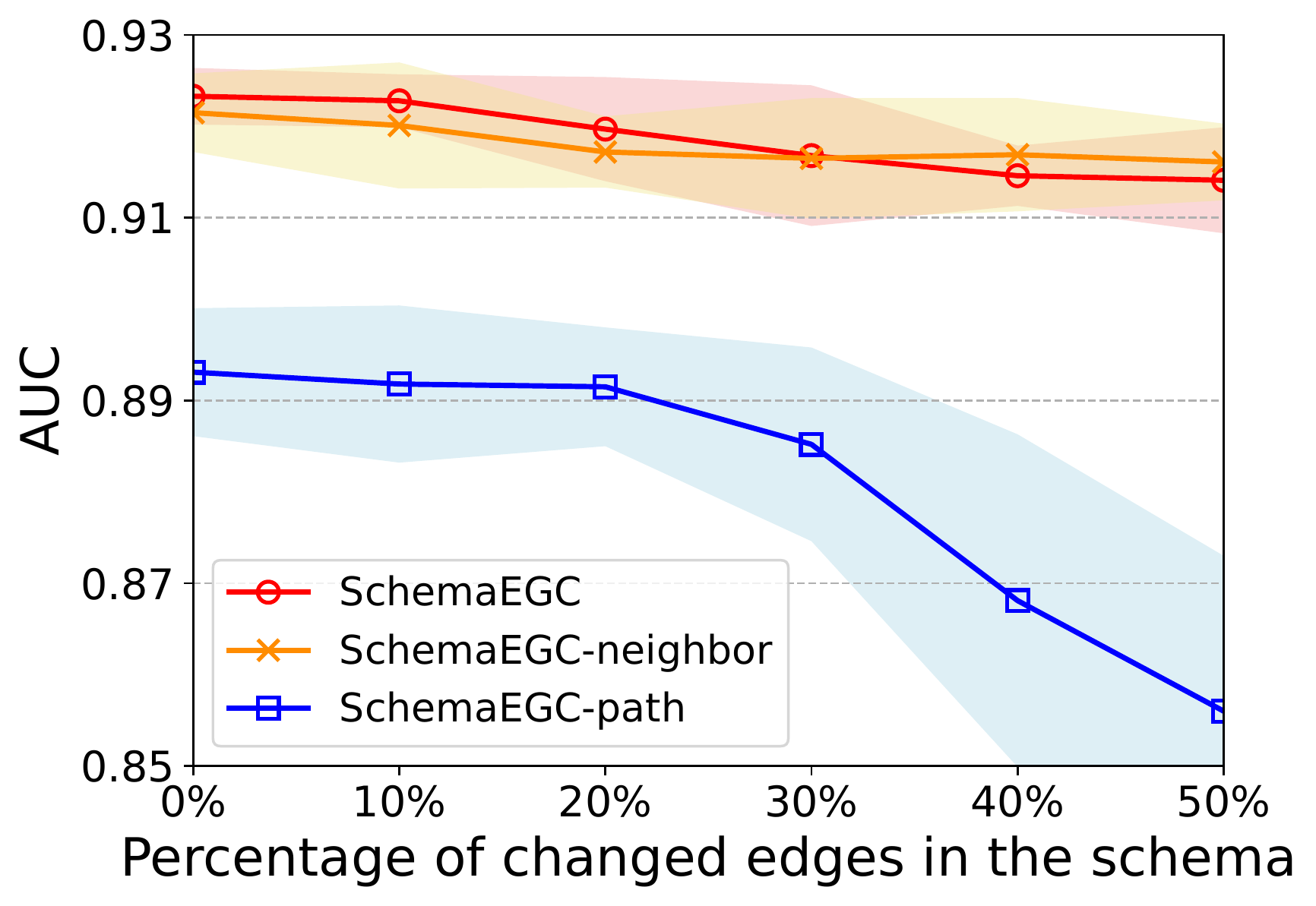}
            \vspace{-0.2in}
            \caption{Impact of schema quality on Car-Bombings dataset.}
            \label{fig:schema_quality}
        \end{minipage}
        \hfill
        \begin{minipage}[t]{0.69\textwidth}
            \vspace{-1.35in}
            \includegraphics[width=\textwidth]{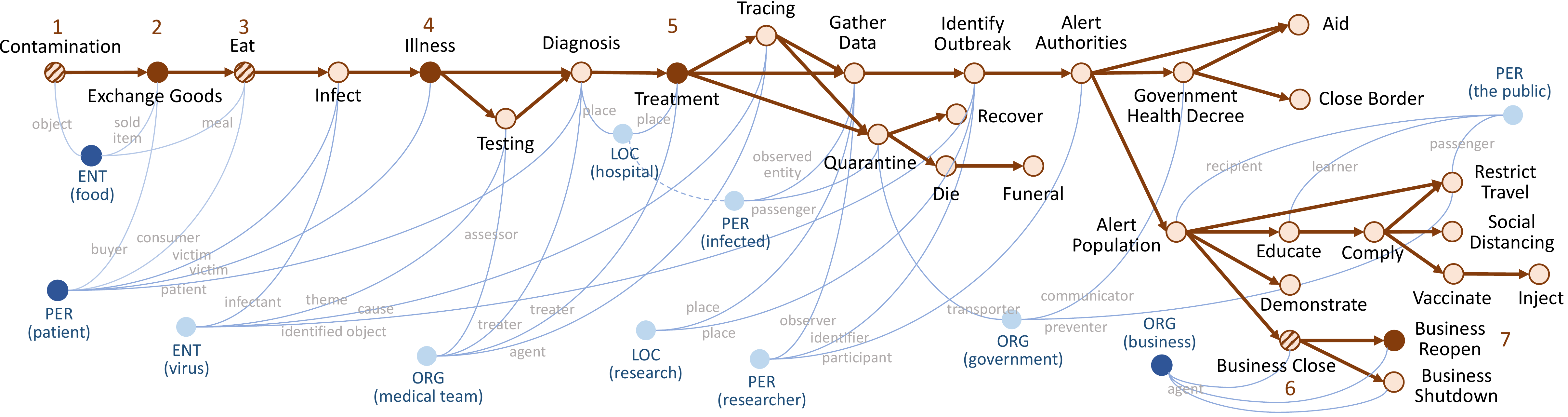}
            \vspace{-0.25in}
            \caption{The Disease-Outbreak schema. Dark red nodes and dark blue nodes denote instantiated events and entities, respectively, and striped nodes are missing events predicted by our model. See Table \ref{table:qualitative} for more detailed description.}
            \label{fig:cs_1}
        \end{minipage}
    \end{figure*}

        We conduct a case study on the predicted result of a test instance graph in Pandemic dataset, which describes the disease outbreak in a Chipotle restaurant.
        Figure \ref{fig:cs_1} demonstrates the key part of the Disease-Outbreak schema, where dark red nodes and dark blue nodes denote instantiated events and entities (i.e., the mapped subgraph) according to the given instance graph, and striped nodes are missing events predicted by our model.
        The detailed description of the instantiated events and the predicted missing events is presented in Table \ref{table:qualitative}.
        Our model can not only predict missing events but also provide the evidence for the prediction by analyzing the attention scores in the neighbor module and the path weights in the path module.
        For example, our model predicts that the No. 3 \textsc{Eat} event is missing from the instance graph, since it is close to two instantiated events: \textsc{Exchange Goods} and \textsc{Illness} (evidence of neighbors), and it can be linked to the mapped subgraph through a couple of high-weight paths such as $(\textsc{TEMP\_REV})$, $(\textsc{meal}, \textsc{sold item})$, and $(\textsc{TEMP}, \textsc{TEMP})$ (evidence of paths).
        Both the evidence of neighbors and the evidence of paths can be used to provide explainability for the predicted results.

    \begin{table*}[t]
	    \centering
	    \small
	    \centering
	    \setlength{\tabcolsep}{3pt}
	    \begin{tabular}{c|c|c|c|c|c}
		    \hline
		    \hline
		    ID & Event type & Description & Event arguments & \makecell[c]{Evidence\\of neighbors} & \makecell[c]{Evidence of paths} \\
		    \hline
		    1 & \textit{Contamination} & \makecell[l]{The three chicken tacos are contaminated\\in the restaurant}  & \makecell[l]{Thing:  Three chicken tacos} & 2 & \makecell[c]{(TEMP)\\(object, sold item)\\(object, meal, TEMP\_REV)} \\
		    \hline
		    2 & \textbf{\makecell[c]{Exchange\\Goods}} & \makecell[l]{Filip Syzller purchased three chicken tacos\\at the Powell Chipotle restaurant} & \makecell[l]{Buyer:  Filip Syzller\\Sold Item:  Three chicken tacos\\Location:  Powell Chipotle} & - & - \\
		    \hline
		    3 & \textit{Eat} & \makecell[l]{Filip Syzller ate the chicken tacos that are\\contaminated}  & \makecell[l]{Consumer: Filip Syzller \\ Meal: Three chicken tacos} & 2, 4 & \makecell[c]{(TEMP\_REV)\\(meal, sold item)\\(TEMP, TEMP)\\(consumer, victim)\\(consumer, patient, TEMP)} \\
		    \hline
		    4 & \textbf{Illness} & \makecell[l]{Filip Syzller became ill with nausea, headaches,\\and severe abdominal pain in Ohio} & \makecell[l]{Patient: Filip Syzller\\Disease: Nausea/Headaches\\Location: Ohio} & - & - \\
		    \hline
		    5 & \textbf{Treatment} & \makecell[l]{Filip Syzller was cured via medical treatment}  & \makecell[l]{Patient: Filip Syzller} & - & - \\
		    \hline
		    6 & \textit{Business Close} & \makecell[l]{Powell Chiptole was closed due to the disease\\outbreak}  & \makecell[l]{Agent: Powell Chipotle} & 7 & \makecell[c]{(TEMP\_REV)\\(agent, agent)} \\
		    \hline
		    7 & \textbf{Business Reopen} & \makecell[l]{Powell Chipotle reopened its restaurant}  & \makecell[l]{Agent: Powell Chipotle} & - & - \\
		    \hline
		    8 & Investigate & \makecell[l]{Chipotle Mexican Grill, Inc. investigated\\Powell Chipotle}  & \makecell[l]{Location: Powell Chipotle\\Examiner: Chipotle Mexican Grill} & - & - \\
		    \hline
		    9 & Announcement & \makecell[l]{The Delaware General Health Department\\cited Powell Chipotle for not maintaining\\the proper temperature for some food items}  & \makecell[l]{Announcer:  Health Department\\Hearer:  Powell Chipotle\\Location:  Powell} & - & - \\
		    \hline
		    \hline
    	\end{tabular}
    	\vspace{0.05in}
		\caption{Case study on the predicted result for a complex event regarding the disease outbreak in a Chipotle restaurant. The bold events means that they are instantiated in the schema graph, the italic events means that they are the missing events predicted by our model, while the last two events cannot be matched to the schema. Evidence of neighbors and evidence of paths denote the neighbors and paths that are important for our model to make the prediction, respectively.}
	\label{table:qualitative}
	\vspace{-0.1in}
\end{table*}

\section{Related Work}
    In this section, we discuss three lines of related work: event schema induction, knowledge graph completion, and graph matching.
    
    \xhdr{Event Schema Induction}
    Event schema induction aims to automatically learn and induce event schemas from event instances.
    The first class of event schema induction methods are sequence-based, which takes event-event relations into account, and orders event structures into sequences \cite{granroth2016happens,weber2018event,weber2020causal}.
    Since they fail to capture the multi-dimensional evolution of real-world complex events, i.e., an event can be preceded or followed by multiple events, researchers propose graph-based schema induction methods, which use graphs to formulate event schemas \cite{wanzare2016crowdsourced,li2020connecting,li2021future}.
    For example, \citet{li2021future} train an auto-regressive graph generation model on instance event graphs and then generate the event schema event by event.
    Our work differs from these methods in that their goal is to learn event schemas, while we focus on the downstream task of event schema, i.e., event graph completion.
    Nonetheless, these methods can be used to output event schemas for our schema-guided model.
    
    \xhdr{Knowledge Graph Completion}
    Event graphs are a type of heterogeneous graphs, which is conceptually related to traditional entity-centric knowledge graphs.
    Many knowledge graph completion methods are embedding-based \cite{bordes2013translating,yang2015embedding,sun2018rotate,kazemi2018simple,zhang2019quaternion}, which learn an embedding vector for each entity and relation by minimizing a predefined loss function on all triplets.
    Such methods have the advantage that they consider the structural context of a given entity in the KG but they fail to capture the multiple relationships (paths) between the head and the tail entity.
    In contrast, the second class of methods is rule-based \cite{galarraga2015fast,sadeghian2019drum,yang2017differentiable}, which aims to learn general logical rules from knowledge graphs by modeling paths between the head and the tail entities.
    However, a significant drawback of these methods is that meaningful rules are usually very sparse, which limits their capability of predicting missing relations that are not covered by known rules.
    Similar to \citet{wang2021relational}, our method can be seen as combining the methodology of the two classes of methods, but our method is specifically designed for event graphs.
    
    \xhdr{Graph Matching}
    Due to the NP-hardness of exact graph matching algorithms \cite{conte2004thirty}, neural graph matching is proposed to solve the problem by machine learning and deep neural networks
    \cite{cho2014finding,li2019graph,petric2019got,lou2020neural}. 
    For example, Graph Matching Networks \citep{li2019graph} learns the node matching relations between two graphs by an attention-based graph neural networks.
    However, the graphs in their study and ours are quite different, because the nodes in event graphs have type information and they should be strictly matched, which will greatly reduce the matching space.
    Therefore, our proposed subgraph matching algorithm fully utilizes the event type information and runs much faster than traditional or neural network based methods.

\section{Conclusion and Future Work}
    We propose a schema-guided method for event graph completion, which overcomes the drawbacks of existing graph completion methods and enables the model to predict new events that are missing in the instance event graphs.
    We consider two factors when modeling the event schema graph, i.e. neighbors and paths, to fully capture its high-order topological and semantic information.
    Experimental results on four datasets and three schemas demonstrate that our method achieves state-of-the-art performance on event graph completion task.
    Moreover, it is resistant to noise in the schema and exhibits high explainability for the prediction results.
    
    Note that in this work, we focus on predicting missing nodes for event graphs.
    Since the links in event graphs could also be noisy, a future direction is therefore to refine links in event graphs based on event schemas.
    In addition, in some event schemas, events are organized in a hierarchical manner (e.g., a pandemic event may include multiple episodes such as \textsc{Authority Response} and \textsc{Society Response}, and each episode may include multiple atomic events, e.g., \textsc{Authority Response} includes \textsc{Restrict Travel} and \textsc{Close Border}).
    We plan to utilize such hierarchy to improve the model performance.
\begin{figure*}[htbp]
		\centering
        \begin{subfigure}[b]{0.3\textwidth}
            \includegraphics[width=\textwidth]{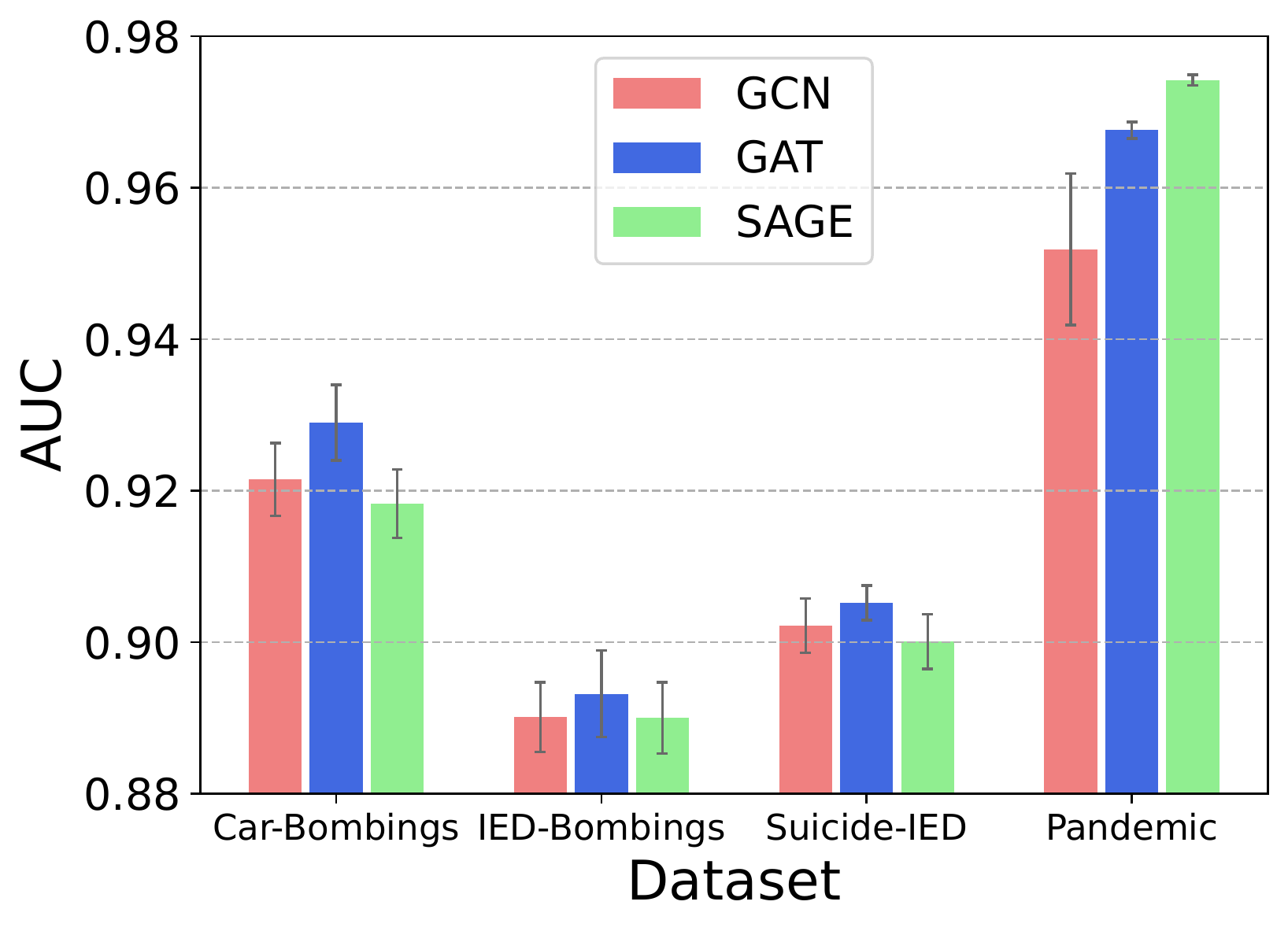}
            \caption{GNN type}
            \label{fig:ps_gnn}
        \end{subfigure}
        \hfill
        \begin{subfigure}[b]{0.3\textwidth}
            \includegraphics[width=\textwidth]{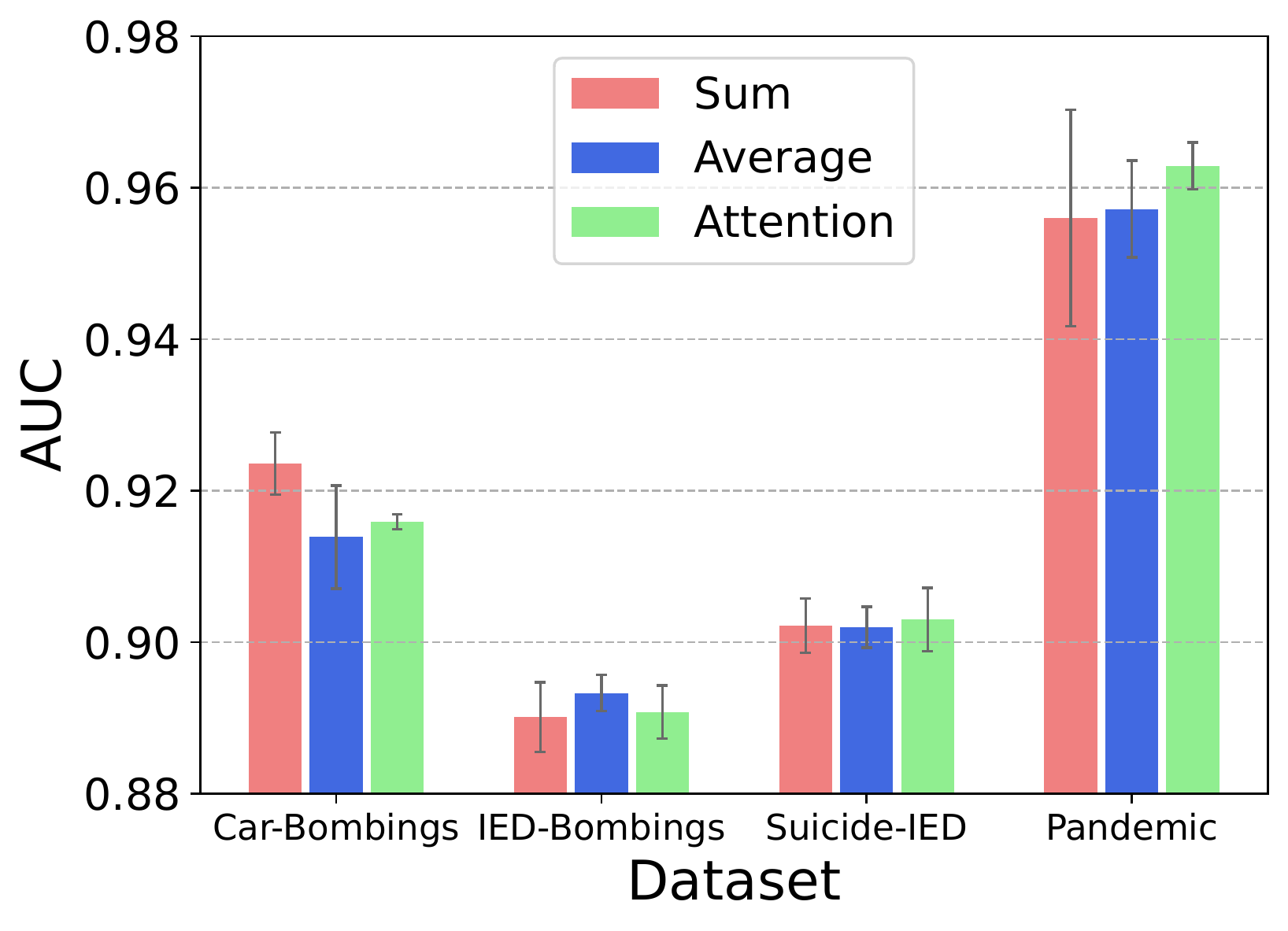}
            \caption{READOUT function}
            \label{fig:ps_pooling}
        \end{subfigure}
        \hfill
        \begin{subfigure}[b]{0.3\textwidth}
            \includegraphics[width=\textwidth]{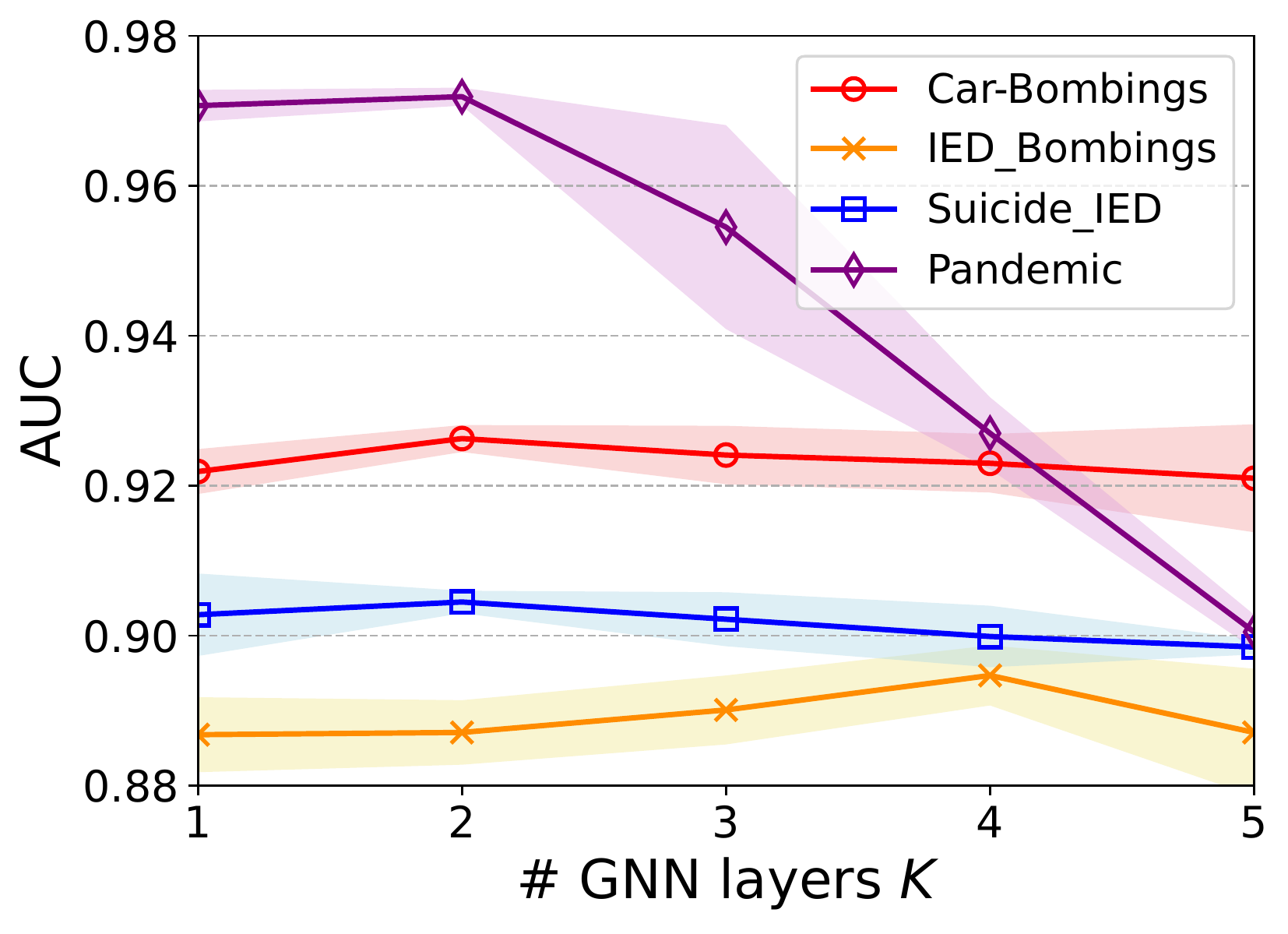}
            \caption{\# GNN layers $K$}
            \label{fig:ps_layer}
        \end{subfigure}
        \vspace{-0.15in}
        \caption{Hyperparameter sensitivity of \alg-Neighbor w.r.t. GNN type, READOUT function, and the number of GNN layers $K$.}
    \end{figure*}
\begin{figure}[htbp]
        \centering
        \begin{subfigure}[b]{0.235\textwidth}
                \includegraphics[width=\textwidth]{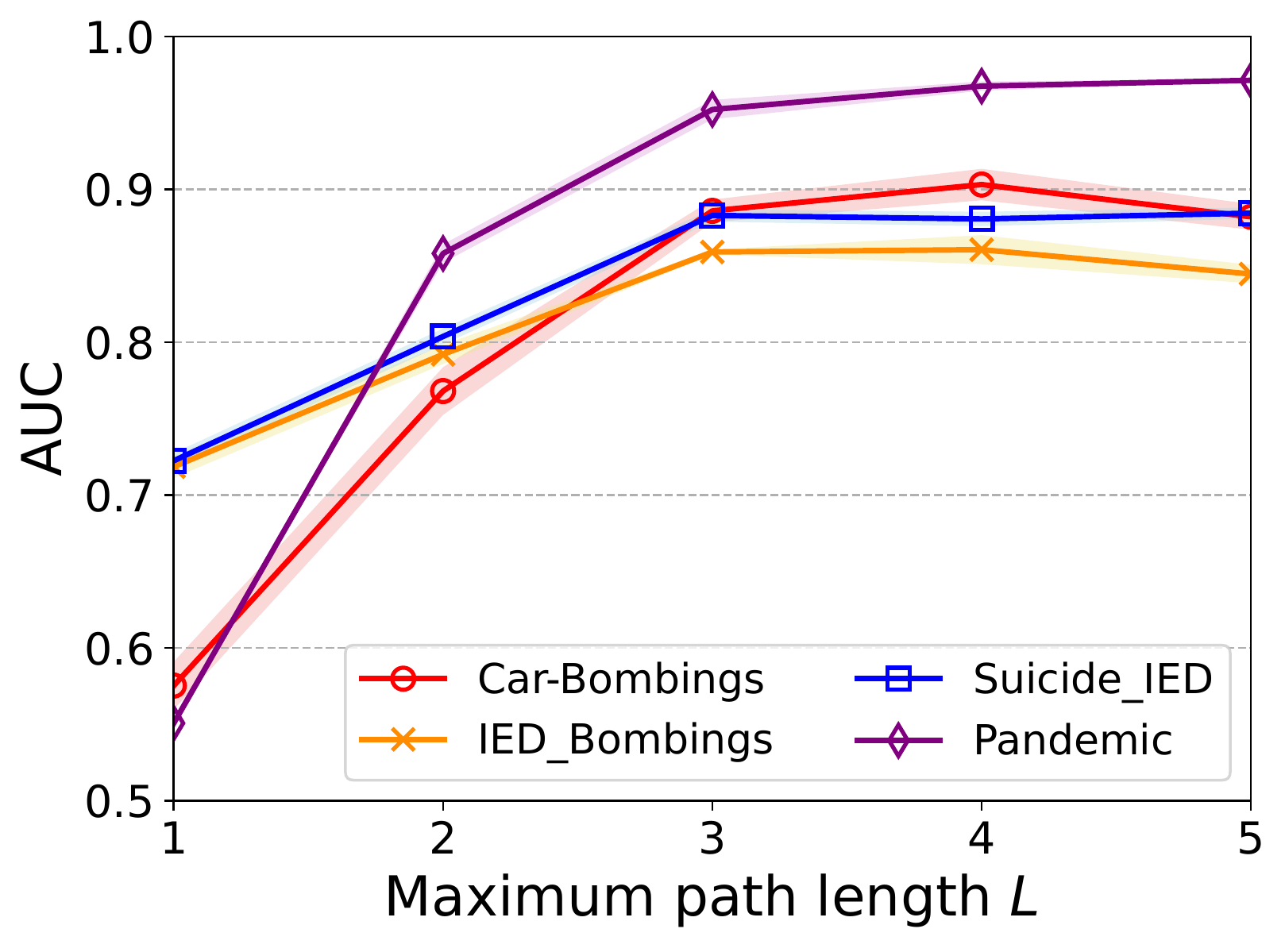}
                \caption{Maximum path length $L$}
                \label{fig:ps_len}
            \end{subfigure}
            \hfill
            \begin{subfigure}[b]{0.235\textwidth}
                \includegraphics[width=\textwidth]{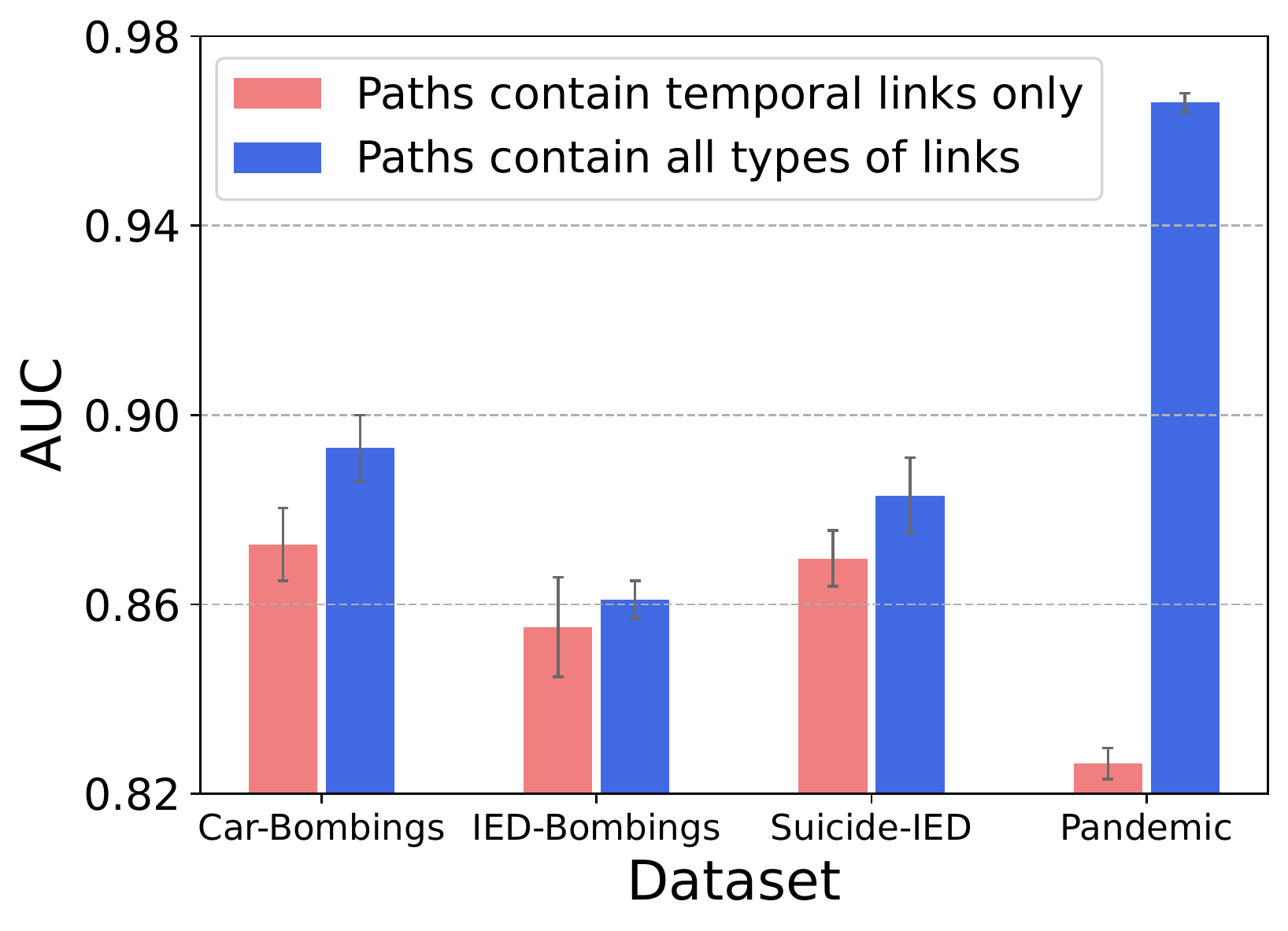}
                \caption{Path type}
                \label{fig:ps_path_event_only}
            \end{subfigure}
            \vspace{-0.2in}
            \caption{Hyperparameter sensitivity of \alg-Path w.r.t. the maximum path length $L$ and path type.}
    \end{figure}
\section{Acknowledgement}
This research is based upon work supported in part by U.S. DARPA KAIROS Program No. FA8750-19-2-1004. The views and conclusions contained herein are those of the authors and should not be interpreted as necessarily representing the official policies, either expressed or implied, of DARPA, or the U.S. Government. The U.S. Government is authorized to reproduce and distribute reprints for governmental purposes notwithstanding any copyright annotation therein.
\bibliographystyle{ACM-Reference-Format}
\bibliography{reference}

\renewcommand\thesubsection{\Alph{subsection}}

\section*{Appendix}
    \subsection{Implementation Details}
        \label{app:implementation}
        \xhdr{Baseline Methods}
        For ID-MLP and Type-MLP, we use an MLP with one hidden layer of $100$ units as the prediction model.
        For TransE and RotatE, we first use them to learn the embedding of each node in the schema graph, then average the embeddings of nodes in a subgraph as the subgraph embedding.
        For an input pair $(e, I')$, their embeddings are concatenated, followed by an MLP with one hidden layer of $100$ units to predict the missing probability.
        The code of TransE and RotatE is from \url{https://github.com/DeepGraphLearning/KnowledgeGraphEmbedding}.
        We set the dimension of entity embedding to $256$ and keep other hyperparameters as default.
        
        \xhdr{Our Method}
        Our method is implemented in PyTorch \cite{paszke2019pytorch}.
        We use GCN \cite{kipf2017semi} as the implementation of GNN in the neighbor module.
        The number of GNN layers $K$ is $3$, and the dimension of hidden layers is $256$.
        We use sum as the READOUT function.
        The maximum path length $L$ is $4$ in the path module.
        ${\rm MLP}_{neighbor}$ and ${\rm MLP}_{path}$ are both MLPs with one hidden layer of dimension $256$.
        We train the model for $20$ epochs with a batch size of $128$, using Adam \cite{kingma2015adam} optimizer with a learning rate of $0.005$.
        The above hyperparameters are determined by maximizing the AUC on the validation set of Car-Bombings, and kept unchanged for all dataset. 
        The result of hyperparameter sensitivity is provided in Appendix \ref{app:hs}.
        The search space of hyperparameters are as follows:
        \begin{itemize}
            \item GNN type: \{GCN, GAT, SAGE\};
            \item The number of GNN layers: $\{1, 2, 3, 4, 5\}$;
            \item The dimension of GNN hidden layers: $\{32, 64, 128, 256, 512\}$;
            \item READOUT function: \{sum, average, attention\};
            \item The maximum path length: $\{1, 2, 3, 4, 5\}$;
            \item The dimension of hidden layer in ${\rm MLP}_{neighbor}$ and ${\rm MLP}_{path}$: $\{32, 64, 128, 256, 512\}$;
            \item Batch size: $\{32, 64, 128, 256, 512\}$;
            \item Learning rate: $\{0.0005, 0.001, 0.005, 0.01, 0.05, 0.1\}$.
        \end{itemize}

    \subsection{Hyperparameter Sensitivity}
        \label{app:hs}
        We study the sensitivity of our model to several key hyperparameters to provide better understanding on the proposed model and investigate its robustness.
            
            For the neighbor module of \alg, we study the impact of GNN type, READOUT function, and the number of GNN layers $K$ on the model performance.
            As shown in Figure \ref{fig:ps_gnn}, we use three GNNs in the neighbor module: GCN \cite{kipf2017semi}, GAT \cite{velivckovic2018graph}, and SAGE \cite{hamilton2017inductive}.
            The number of GNN layers is $3$, and the dimension of hidden layers is $256$ (for GAT, the number of attention heads is $16$ and the dimension of each attention head is $16$).
            The result shows that GAT achieves the best performance in the three IED-related scenarios while SAGE performs the best in the pandemic scenario.
            Figure \ref{fig:ps_pooling} demonstrates that attention is a good choice for the READOUT function, but it does not perform universally the best on all datasets.
            Figure \ref{fig:ps_layer} shows that our model usually achieves the best performance when the number of GNN layers is $2 \sim 4$.
            Moreover, Pandemic dataset is much more sensitive to the number of GNN layers.
            
            For the path module of SchemaEGC, we study the impact of the maximum path length and the path type.
            From Figure \ref{fig:ps_len} it is clear that our model is extremely sensitive to the maximum path length $L$, since the AUC increases by $0.2 \sim 0.4$ as $L$ goes from $1$ to $5$.
            But the marginal improvement is diminishing when $L \geq 3$ due to the issue of overfitting.
            Figure \ref{fig:ps_path_event_only} demonstrates that considering all the three types of links for the path module works better than only considering event-event temporal links.

\end{document}